\documentclass{article}
\usepackage{spconf,amsmath,amssymb,graphicx,color,colortbl}
\definecolor{lightgray}{gray}{0.85}

\def\boldb{{\mathbf b}}
\def\h{{\mathbf h}}

\def\R{\mathbb{R}}

\DeclareMathOperator*{\argmax}{arg\,max}
\DeclareMathOperator*{\argmin}{arg\,min}

\title{On the Compression of Recurrent Neural Networks with an Application to
{LVCSR} acoustic modeling for Embedded Speech Recognition}
\name{Rohit Prabhavalkar$^\dagger$ \qquad Ouais Alsharif~$^\dagger$ \qquad Antoine Bruguier \qquad
Ian McGraw\thanks{$^\dagger$Equal contribution. The authors would like to
    thank Ha{\c{s}}im Sak and Raziel Alvarez for helpful comments and
    suggestions on this work,
and Chris Thornton and Yu-hsin Chen for comments on an earlier draft.}}
\address{
  Google Inc.\\
  \{prabhavalkar,oalsha,tonybruguier,imcgraw\}@google.com
}
\begin{document}
%
\maketitle
\begin{abstract}
  We study the problem of compressing recurrent neural networks (RNNs). In
particular, we focus on the compression of RNN acoustic models, which are
motivated by the goal of building compact and accurate speech recognition
systems which can be run efficiently on mobile devices. In this work, we present
a technique for general recurrent model compression that jointly compresses both
recurrent and non-recurrent inter-layer weight matrices. We find that the
proposed technique allows us to reduce the size of our Long Short-Term Memory
(LSTM) acoustic model to a third of its original size with negligible loss in
accuracy.

\end{abstract}
\begin{keywords}
model compression, LSTM, RNN, SVD, embedded speech recognition
\end{keywords}
\section{Introduction}
\label{sec:introduction}
Neural networks (NNs) with multiple
feed-forward~\cite{SeideLiYu11,HintonDengYuEtAl12}
or recurrent hidden layers~\cite{SakSeniorBeaufays14,
SainathVinyalsSeniorEtAl15} have emerged as state-of-the-art acoustic models
(AMs) for automatic speech recognition (ASR) tasks. Advances in computational
capabilities coupled with the availability of large annotated speech corpora
have made it possible to train NN-based AMs with a large number of
parameters~\cite{DengYu14} with great success.

As speech recognition technologies continue to improve, they are becoming
increasingly ubiquitous on mobile devices: voice assistants such as Apple's
Siri, Microsoft's Cortana, Amazon's Alexa and
Google Now~\cite{SchalkwykBeefermanBeaufaysEtAl10}
enable users to search for information using their voice. Although the
traditional model for these applications has been to recognize speech remotely
on large servers, there has been growing interest in developing ASR technologies
that can recognize the input speech directly
``on-device''~\cite{LeiSeniorGruensteinEtAl13}.
This has the promise to reduce latency while
enabling user interaction even in cases where a mobile data connection is
either unavailable, slow or unreliable. Some of the main challenges in this
regard are the disk, memory and computational constraints imposed by these
devices. Since the number of operations in neural networks is proportional to
the number of model parameters,
compressing the model is desirable from the point of view of
reducing memory usage and power consumption.

In this paper, we study techniques for compressing recurrent neural networks
(RNNs), specifically RNN acoustic models. We demonstrate how a generalization of
conventional inter-layer matrix factorization techniques (e.g.,
~\cite{XueLiGong13, XueLiYuEtAl14}), where we
jointly compress both recurrent and inter-layer weight matrices, allows us
to compress acoustic models up to a third of their original size with negligible
loss in accuracy. While we focus on acoustic modeling, the techniques presented
can be applied to RNNs in other domains, e.g., handwriting
recognition~\cite{graves2009novel} and machine
translation~\cite{sutskever2014sequence} inter alia. The technique presented
in this paper encompasses both traditional recurrent neural
networks (RNNs) \emph{as well as} Long Short-Term Memory (LSTM) neural networks.

In Section~\ref{sec:related-work}, we review previous work that has focussed on
techniques for compressing neural networks. Our proposed compression technique
is presented in Section~\ref{sec:compression}. We examine the effectiveness of
proposed techniques in Sections~\ref{sec:experimental-setup}
and~\ref{sec:results}. Finally, we conclude with a discussion of our findings in
Section~\ref{sec:conclusions}.

\section{Related Work}
\label{sec:related-work}
There have been a number of previous proposals to compress neural networks, both
in the context of ASR as well as in the broader field of machine learning.
We summarize a number of proposed approaches in this section.

It has been noted in previous work that there is a large amount of redundancy in
the parameters of a neural network. For example, Denil et
al.~\cite{DenilShakibiDinhEtAl13} show that the entire neural network can be
reconstructed given the values of a
small number of parameters. Caruana and colleagues show that the output
distribution learned by a larger neural network can be approximated by a neural
network with
fewer parameters by training the smaller network to directly predict the outputs
of the larger network~\cite{BuciluaCaruanaNiculescu-Mizil06, BaCaruana14}.
This approach, termed ``model
compression''~\cite{BuciluaCaruanaNiculescu-Mizil06} is closely related to the
recent ``distillation" approach proposed by Hinton
et al.~\cite{HintonVinyalsDean15}. The redundancy in a neural network has also
been exploited in the HashNet approach of Chen et
al.~\cite{ChenWilsonTyreeEtAl15}, which imposes parameter tying in network based
on a set of hash functions.

In the context of ASR, previous approaches to acoustic model compression have
focused mainly on the case of feed-forward DNNs. One popular technique is based
on sparsifying the weight matrices in the neural network, for example, by
setting weights whose magnitude falls below a certain threshold to
zero~\cite{SeideLiYu11} or based on the second-derivative of the loss function
in the ``optimal brain damage'' procedure~\cite{LeCunDenkerSolla89}. In fact,
Seide et al.~\cite{SeideLiYu11} demonstrate that up to two-thirds of the weights
of the feed-forward network can be set to zero without incurring any loss in
performance. Although techniques based on sparsification do decrease the number
of effective weights, encoding the subset
of weights which can be `zeroed out' requires additional memory.
Further, if the weight matrices are represented as dense matrices for
efficient computation, then the parameter savings on disk will not translate in
to savings of runtime memory. Other techniques to reduce the
number of model parameters is based on changing the neural network
architecture, e.g., by introducing bottleneck layers~\cite{GrezlFousek08}
or through a low-rank matrix factorization
layer~\cite{SainathKingsburySindhwaniEtAl13}. We also note recent work by Wang
et al.~\cite{WangLiGong2015} which uses a combination of singular value
decomposition (SVD) and vector quantization to compress acoustic models.

The methods investigated in our work are most similar to previous work that
has examined using SVD to reduce the number of
parameters in the network in the context of feed-forward
DNNs~\cite{XueLiGong13, XueLiYuEtAl14, NakkiranAlvarezPrabhavalkarEtAl15}.
As we describe in Section~\ref{sec:compression}, our methods can be thought of
as an extension of the techniques proposed by Xue et al.~\cite{XueLiGong13},
wherein we jointly factorize both recurrent and (non-recurrent) inter-layer
weight matrices in the network.


\section{Model Compression}
\label{sec:compression}
In this section, we present a general technique for compressing individual
recurrent layers in a recurrent neural network, thus generalizing the methods
proposed by Xue et al.~\cite{XueLiGong13}.

We describe our approach in the most general setting of a standard RNN. We
denote the activations of the $l$-th hidden layer, consisting of $N^l$ nodes,
at time $t$ by $\h^{l}_t \in \R^{N^l}$. The inputs to this layer at time $t$ --
which are in turn the activations from the
previous layer or the input features -- are denoted by $\h^{l-1}_t \in
\R^{N^{l-1}}$. We can then write the following equations which define the
output activations of the $l$-th
and $(l+1)$-th layers in a standard RNN:
\begin{align}
  \label{eq:basic-eqns1}
\h^{l}_t     &=& &\sigma ( W^{l-1}_x \h^{l-1}_t & &+& & W^{l}_h \h^{l}_{t-1} & &+ \boldb^{l}) & \\
  \label{eq:basic-eqns2}
\h^{l+1}_t &=& &\sigma ( W^{l}_x    \h^{l}_t    &  &+& &  W^{l+1}_h \h^{l+1}_{t-1} & &+ \boldb^{l+1} ) &
\end{align}
\noindent where, $b^{l} \in \R^{N^l}$ and $b^{l+1} \in \R^{N^{l+1}}$
represent bias vectors, $\sigma(\cdot)$ denotes a non-linear activation
function, and $W^{l}_{x} \in \R^{N^{l+1} \times N^{l}}$ and
$W^{l}_{h} \in \R^{N^l \times N^l}$ denote weight matrices that we refer
to respectively as the \emph{inter-layer} and the \emph{recurrent} weight
matrices, respectively\footnote{The
equations are slightly more complicated when using LSTM cells in the recurrent
layer, but the basic form remains the same. See
Section~\ref{sec:lstm-compression}.}. Since our proposed approach can be
applied independently for each recurrent hidden layer, we only describe the
compression operations for a particular layer $l$. We jointly compress the
recurrent and inter-layer matrices corresponding to a specific layer $l$ by
determining a
suitable recurrent projection matrix~\cite{SakSeniorBeaufays14}, denoted by
$P^l \in \R^{r^l \times N^l}$, of rank $r^{l} < N^l$ such
that, $W^l_h = Z^l_h P^l$ and $W^l_x = Z^l_x P^l$, thus allowing us to
re-write~\eqref{eq:basic-eqns1} and~\eqref{eq:basic-eqns2} as,
\begin{align}
  \h^{l}_t    &=& &\sigma ( W^{l-1}_x \h^{l-1}_t & &+& & Z^{l}_h P^{l} \h^{l}_{t-1} & &+ \boldb^{l})& \\
  \h^{l+1}_t &=& &\sigma ( Z^{l}_x P^{l} \h^{l}_t &   &+& & W^{l+1}_h \h^{l+1}_{t-1} & &+ \boldb^{l+1} ) &
\end{align}
\begin{figure}
    \centering
    \includegraphics[width=\columnwidth]{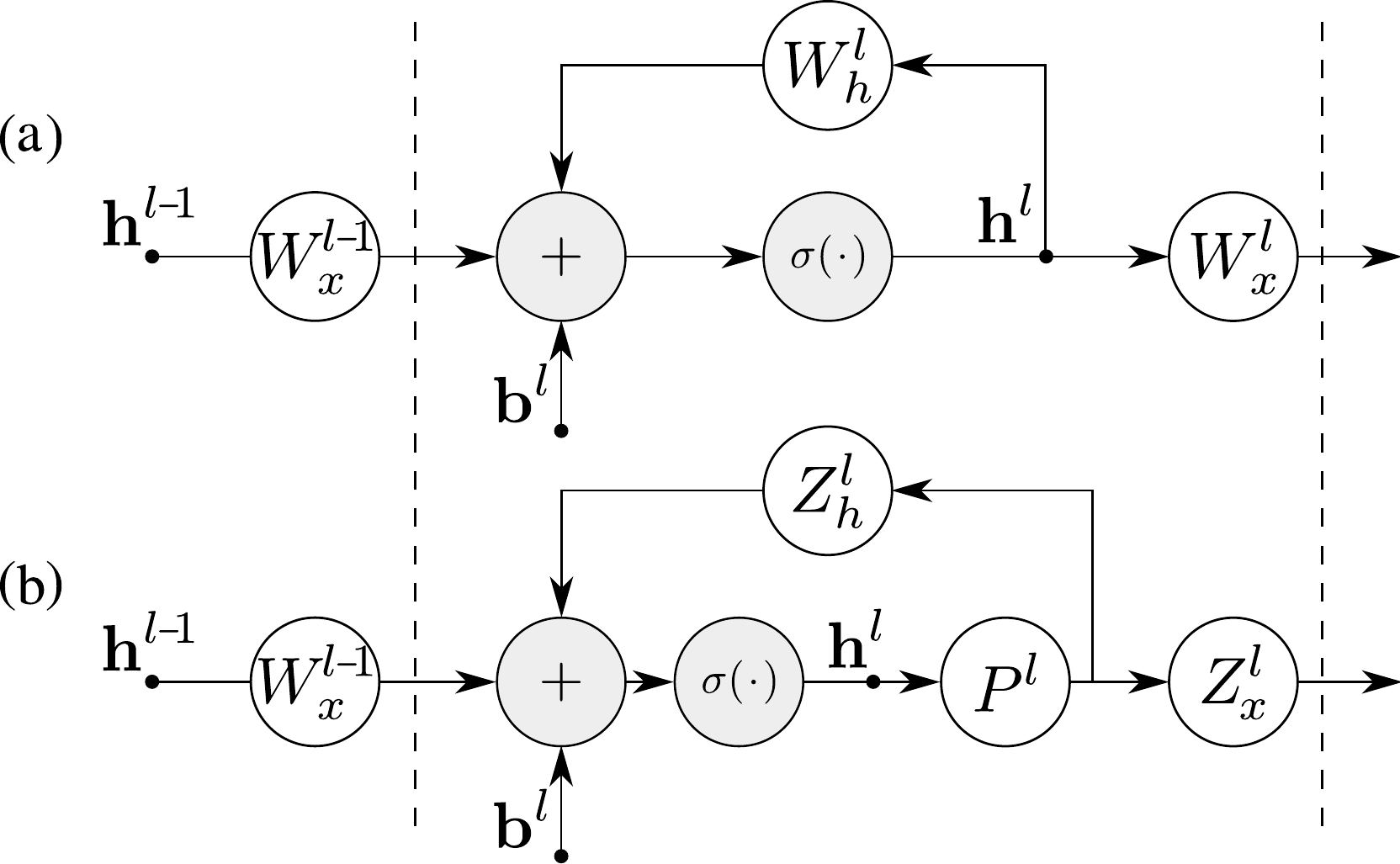}
    \caption{The initial model (Figure (a)) is compressed by jointly factorizing
      recurrent ($W^l_h$) and inter-layer ($W^l_x$) matrices, using a shared
    recurrent projection matrix ($P^l$)~\cite{SakSeniorBeaufays14} (Figure (b)).}
  \label{fig:compression}
\end{figure}
\noindent where, $Z^l_h \in \R^{N^l \times r^l}$ and
$Z^l_x \in \R^{N^{l+1} \times r^l}$. This compression process is depicted
graphically in Figure~\ref{fig:compression}.

We note that sharing $P^l$ across the recurrent and inter-layer matrices allows
for more efficient parameterization of the weight matrices; as shown in
Section~\ref{sec:results}, this does not result in a significant loss of
performance. Thus, the degree of compression in the model can be controlled by
setting the ranks $r^{l}$ of the projection matrices in each of the layers of
the network.

We determine
the recurrent projection matrix $P^{l}$, by first computing an SVD of the
recurrent weight matrix, which we then truncate, retaining only the top
$r^l$ singular values (denoted by $\widetilde{\Sigma^{l}_h}$) and the
corresponding singular vectors from $U^{l}_h$ and $V^{l}_h$ (denoted by
$\widetilde{U^{l}_h}$ and $\widetilde{V^{l}_h}$, respectively):
\begin{align}
\label{eqn:rank-truncate}
W^{l}_h = U^{l}_h \Sigma^{l}_h {V^{l}_h}^T \approx \left( \widetilde{U^{l}_h} \widetilde{\Sigma^{l}_h} \right) \widetilde{V^{l}_h}^T = Z^l_h P^l
\end{align}
\noindent where $Z^l_h = \widetilde{U^{l}_h} \widetilde{\Sigma^{l}_h}$ and
$P^{l} = \widetilde{V^{l}_h}^T$. Finally, we determine $Z^{l}_x$, as the
solution to the following least-squares problem:
\begin{equation}
  Z^l_x = \argmin_{Y} \| Y P^l - W^l_x \|^2_{\mathcal{F}}
\end{equation}
\noindent where, $\|X\|_{\mathcal{F}}$ denotes the Frobenius norm of the matrix.
In pilot experiments we found that the proposed SVD-based initialization
performed better than training a model with recurrent projection matrices
(i.e., same model architecture) but with random initialization of the network
weights.

\subsection{Applying our technique to LSTM RNNs}
\label{sec:lstm-compression}
Generalizing the procedure described above in the context of \emph{standard} RNNs to
the case of LSTM RNNs~\cite{SakSeniorBeaufays14, SakSeniorRaoEtAl15a,
SakSeniorRaoEtAl15b} is straightforward. Using the notation
in~\cite{SakSeniorBeaufays14}, note that the
recurrent-weight matrix $W^{l}_h$ in the case of the LSTM is the concatenation
of the four \emph{gate weight matrices}, obtained by stacking them vertically:
$$ \left[ W_{im}, W_{om}, W_{fm}, W_{cm} \right]^T $$ which represent
respectively, recurrent connections to the input gate, the output gate, the
forget gate and
the cell state. Similarly, the inter-layer matrix $W^{l}_x$ is the concatenation
of the matrices:
$$ \left[ W_{ix}, W_{fx}, W_{ox}, W_{cx} \right]^T $$ which
correspond to the input gate, the forget gate, the output gate and the cell
state (of the next layer). With these definitions, compression can be applied as described in
Section~\ref{sec:compression}. Note that we do not compress the ``peep-hole"
weights, since they are already narrow, single column matrices and do not
contribute significantly to the total number of parameters in the network.



\section{Experimental Setup}
\label{sec:experimental-setup}

In order to determine the effectiveness of the proposed RNN compression
technique, we conduct experiments on a open-ended large-vocabulary dictation
task.

As we mentioned in Section~\ref{sec:introduction}, one of our primary
motivations behind investigating acoustic model compression is to build compact
acoustic models that can be deployed on mobile devices. In recent work, Sak et
al. have demonstrated that deep LSTM-based AMs trained to predict either
context-independent (CI) phoneme targets~\cite{SakSeniorRaoEtAl15a} or
context-dependent (CD) phoneme targets~\cite{SakSeniorRaoEtAl15b} approach
state-of-the-art performance on speech tasks. These systems have two important
characteristics: in addition to the CI or CD phoneme labels, the system can
also hypothesize a ``blank'' label if it is unsure of the identity of the
current phoneme, and the systems are trained to optimize the connectionist
temporal classification (CTC) criterion~\cite{GravesFernandesGomezEtAl06} which
maximizes the total probability of correct label sequence conditioned on the
input sequence. More details can be found in~\cite{SakSeniorRaoEtAl15a,
SakSeniorRaoEtAl15b}.

Following~\cite{SakSeniorRaoEtAl15a}, our baseline model is thus a
\emph{CTC model}: a five hidden layer RNN with 500 LSTM cells in each layer,
which predicts 41 CI phonemes (plus ``blank''). As a point of comparison, we
also present results obtained using a much larger state-of-the-art
`server-sized' model which is too large to deploy on embedded
devices but nonethless serves as an upper-bound performance for our models
on this dataset. This model consists of five hidden layers with 600 LSTM cells
per layer, and is trained to predict one of 9287 context-dependent phonemes
(plus ``blank'').

Our systems are trained using distributed asynchronous stochastic gradient
descent with a parameter server~\cite{DeanCorradoMongaEtAl12}. The
systems are first trained to convergence to optimize the CTC criterion,
following which these are discriminatively sequence trained to optimize the
state-level minimum Bayes risk (sMBR)
criterion~\cite{Kingsbury09, SakVinyalsHeigold14}. As
discussed in Section~\ref{sec:results}, after applying the proposed compression
scheme, we further fine-tune the network: first with the CTC criterion, followed
by sequence discriminative training with the sMBR criterion. This additional
fine-tuning step was found to be necessary to achieve good performance,
particularly as the amount of compression was increased.

The language model used in this work is a 5-gram model trained on $\sim$100M
sentences of in-domain data, with entropy-based pruning applied to reduce the
size of the
LM down to roughly 1.5M n-grams (mainly bigrams) with a 64K vocabulary. Since
our goal is to build a recognizer to run efficiently on mobile devices, we
minimize the size of the decoder graph used for recognition, following the
approach outlined in~\cite{LeiSeniorGruensteinEtAl13}: we perform an additional
pruning step to
generate a much smaller \emph{first-pass} language model (69.5K n-grams; mainly
unigrams), which is composed with
the lexicon transducer to construct the decoder graph. We then perform
on-the-fly rescoring with the larger LM. The resulting models, when compressed
for use on-device, total about 20.3 MB, thus enabling them to be run
many times faster than real-time on recent
mobile devices~\cite{McGrawPrabhavalkarAlvarezEtAl16}.

We parameterize the input acoustics by computing
40-dimensional log mel-filterbank energies over the 8Khz range, which are
computed every 10ms over 25ms
windowed speech segments. The server-sized system uses 80-dimensional
features computed over the same range since this resulted in slightly improved
performance. Following~\cite{SakSeniorRaoEtAl15b}, we stabilize CTC
training by stacking together 8 consecutive speech frames (7 right context
frames); only every third stacked frame is presented as an input to the network.

\subsection{Training and Evaluation Data}
\label{sec:training-data}
Our systems are trained on $\sim$3M hand-transcribed anonymized
utterances extracted from Google voice search traffic ($\sim$2000
hours). We create ``multi-style" training data by synthetically distorting
utterances to simulate background noise and reverberation using a room
simulator with noise samples extracted from YouTube videos and environmental
recordings of everyday events; 20 distorted examples are created for each
utterance in the training set. Systems are additionally adapted using the sMBR
criterion~\cite{Kingsbury09, SakVinyalsHeigold14} on a set of $\sim$1M
anonymized hand-transcribed (in-domain) dictation utterances extracted from
Google traffic, processed to generate ``multi-style" training data as described
above, which improves performance on our dictation task. All results are
reported on a set of 13.3K hand-transcribed anonymized utterances extracted from
Google traffic from an open-ended dictation domain.

\section{Results}
\label{sec:results}
In our experiments, we seek to determine the impact of the proposed
joint SVD-based compression technique on system performance. In particular, we
are interested in determining how system performance varies as a function of the
degree of compression, which is controlled by setting the ranks of the recurrent
projection matrices $r^l$ as described in Section~\ref{sec:compression}.

Notice that since the proposed compression scheme is applied to all hidden
layers of the baseline system, there are numerous settings of the ranks $r^l$
for the projection matrices in each layer which result in the same number of
total parameters
in the compressed network. In order to avoid this ambiguity, we set the various
projection ranks using the following criterion: Given a threshold $\tau$, for
each layer $l$, we set the rank $r^l$ of the corresponding projection matrix
such that it corresponds to retaining a fraction of at most $\tau$ of
the \emph{explained variance} after the truncated SVD of $W_{h}^l$. More
specifically, if the singular values in $\Sigma^l_h$
in~\eqref{eqn:rank-truncate} are sorted in non-increasing order as
$\sigma^l_1 \geq \sigma^l_2 \geq \cdots \geq \sigma^l_{N}$, we set
each $r^l$ as:
\begin{equation}
  r^l = \argmax_{1 \leq k \leq N}
  \left\{ \frac{ \sum_{j=1}^k {\sigma^l_j}^2}{ \sum_{j=1}^{N} {\sigma^l_j}^2}
  \leq \tau \right\}
  \label{eq:projection-ranks}
\end{equation}
Choosing the projection ranks using~\eqref{eq:projection-ranks} allows us to
control the degree of compression, and thus compressed model size by varying a
single parameter, $\tau$. In pilot experiments we found that this scheme
performed better than setting ranks to be equal for all layers (given the same
total parameter budget).
Once the projection ranks $r^l$ have been determined for the various projection
matrices we fine-tune the compressed models by first optimizing
the CTC criterion, followed by sequence training with the sMBR criterion and
adaptation on in-domain data as described in Section~\ref{sec:training-data}.
The results of our experiments are presented in Table~\ref{tbl:results1}.
\begin{table}
  \centering
  \begin{tabular}{| c | c | c | c |}
    \hline
    System & Projection ranks, $r^l$ & Params & WER \\
    \hline
    \hline
    server & - & 20.1M & 11.3 \\
    \hline
    baseline & - & 9.7M & 12.4 \\
    \hline
    \hline
    $\tau= 0.95$ & 350, 375, 395, 405, 410 & 8.6M & 12.3 \\
    \hline
    $\tau= 0.90$ & 270, 305, 335, 345, 350 & 7.2M & 12.5 \\
    \hline
    $\tau= 0.80$ & 175, 215, 245, 260, 265 & 5.4M & 12.5 \\
    \hline
    $\tau= 0.70$ & 120, 150, 180, 195, 200 & 4.1M & 12.6 \\
    \hline
    \rowcolor{lightgray}
    $\tau= 0.60$ & 80, 105, 130, 145, 150 & 3.1M & 12.9 \\
    \hline
    $\tau= 0.50$ & 50, 70, 90, 100, 110 & 2.3M & 13.2 \\
    \hline
    $\tau= 0.40$ & 30, 45, 55, 65, 75 & 1.7M & 14.4 \\
    \hline
  \end{tabular}
  \caption{Word error rates (\%) on the test set as a function of the percentage
    of explained variance retained ($\tau$) after the SVDs of the
    recurrent weight matrices $W^{l}_h$ in the hidden layers of the RNN.}
  \label{tbl:results1}
\end{table}

As can be seen in Table~\ref{tbl:results1}, the baseline system which predicts CI
phoneme targets is only $\sim$10\% relative worse than the larger server-sized
system, although it has half as many parameters. Since the ranks $r^l$ are all
chosen to retain a given fraction of the explained variance in the SVD
operation, we also note that earlier hidden layers in the network appear to
have lower ranks than later layers, since most of the variance is accounted for
by a smaller number of singular values. It can be seen from
Table~\ref{tbl:results1} that word error rates increase as the amount of
compression is increased, although performance of the compressed systems are
close to the baseline for moderate compression ($\tau \geq 0.7$).
Using a value of $\tau = 0.6$, enables the model to be compressed to a third of
its original size, with only a small degradation
in accuracy. However, performance begins to degrade significantly for
$\tau \leq 0.5$.
Future work will consider alternative
techniques for setting the projection ranks $r^l$ in order to examine their
impact on system performance.

\section{Conclusions}
\label{sec:conclusions}
We presented a technique to compress RNNs using a
joint factorization of recurrent and inter-layer weight matrices, generalizing
previous work~\cite{XueLiGong13}. The proposed technique was applied
to the task of compressing LSTM RNN acoustic models for embedded speech
recognition, where we found that we could
compress our baseline acoustic model to a third of its original size with
negligible loss in accuracy. The proposed techniques, in combination with weight
quantization, allow us to build a small and efficient speech recognizer that run
many times faster than real-time on recent mobile
devices~\cite{McGrawPrabhavalkarAlvarezEtAl16}.

\bibliographystyle{IEEEbib}
\bibliography{refs}

\end{document}